\documentclass[letterpaper, 10 pt, conference]{ieeeconf}  

\IEEEoverridecommandlockouts                              

\usepackage{moresize}
\usepackage{amsmath}
\usepackage{adjustbox}
\usepackage{tabularx}
\usepackage{nomencl}
\usepackage{ragged2e}
\usepackage{amssymb}
\usepackage{mathtools}
\usepackage{graphicx}
\usepackage{float}
\usepackage{subcaption}
\usepackage{epstopdf}
\usepackage[tableposition=top]{caption}
\usepackage{array}
\usepackage{url}
\usepackage{algorithm}
\usepackage{algpseudocode}
\usepackage{diagbox}
\usepackage{multirow}
\usepackage[table]{xcolor}
\usepackage[framemethod=TikZ]{mdframed}
\usepackage[procnames]{listings}
\usepackage[hidelinks]{hyperref}
\usepackage{tcolorbox}

\mdfdefinestyle{frameStyle}{%
    roundcorner=5pt,
    backgroundcolor=white}

\algtext*{EndIf}
\algtext*{EndFor}
\algtext*{EndWhile}
\algtext*{EndFunction}

\newcommand{\assign}[0]{$\leftarrow$ }

\newcommand{\algnot}[0]{\textbf{not} }
\newcommand{\algin}[0]{\textbf{in} }

\newcommand{\algtrue}[0]{\textbf{true} }
\newcommand{\algfalse}[0]{\textbf{false} }

\title{\LARGE \bf
Property-Based Testing in Simulation for Verifying\\Robot Action Execution in Tabletop Manipulation
}

\author{Salman Omar Sohail$^{\dagger}$, Alex Mitrevski$^{\dagger\mathsection}$, Nico Hochgeschwender$^{\dagger}$, and Paul G. Pl{\"o}ger$^{\dagger}$
\thanks{$^{*}$This work was supported by the b-it foundation and by the European Union's Horizon 2020 project SESAME (grant agreement No. 101017258)} %
\thanks{$^{\dagger}$The authors are with the Autonomous Systems Group, Department of Computer Science, Hochschule Bonn-Rhein-Sieg, Sankt Augustin, Germany
        {\tt\scriptsize salman.sohail@smail.inf.h-brs.de}, {\tt\scriptsize <aleksandar.mitrevski, nico.hochgeschwender, paul.ploeger>@h-brs.de}}
\thanks{$^{\mathsection}$Corresponding author} %
}

\begin{document}

\maketitle
\thispagestyle{empty}
\pagestyle{empty}


\begin{abstract}
An important prerequisite for the reliability and robustness of a service robot is ensuring the robot's correct behavior when it performs various tasks of interest. Extensive testing is one established approach for ensuring behavioural correctness; this becomes even more important with the integration of learning-based methods into robot software architectures, as there are often no theoretical guarantees about the performance of such methods in varying scenarios. In this paper, we aim towards evaluating the correctness of robot behaviors in tabletop manipulation through automatic generation of simulated test scenarios in which a robot assesses its performance using property-based testing. In particular, key properties of interest for various robot actions are encoded in an action ontology and are then verified and validated within a simulated environment. We evaluate our framework with a Toyota Human Support Robot (HSR) which is tested in a Gazebo simulation. We show that our framework can correctly and consistently identify various failed actions in a variety of randomised tabletop manipulation scenarios, in addition to providing deeper insights into the type and location of failures for each designed property.
\end{abstract}


\section{INTRODUCTION}
\label{sec:introduction}

With the integration of autonomous service robots into industries and households, there is a requirement for increased robot safety and dependability. To fulfill these requirements, robot developers need to validate their systems extensively before deployment. In principle, validation is a challenging problem due to factors such as the environment's unpredictability, the robot's lack of knowledge, hardware and software failures, but also due to the fact that a robot may utilise learning-based components for which formal correctness guarantees are difficult to provide.

One common approach for robot verification and validation is testing directly in the real world, but real-world tests are often complex to set up and perform to an extent that would provide sufficient test coverage of a complete robot system. An alternative to this is simulation-based testing, which provides multiple attractive features, such as comparatively low setup and execution costs, speed, scalability, robot safety, as well as a possibility to automate tests \cite{sotiropoulos_2017}, at the cost of sacrificing the realism of real-world testing. Setting up and executing simulated tests can, however, be a challenging problem on its own. In particular, simulated scenarios for a robot are often hand-crafted and each scenario has to be manually altered when the need arises to test different scenarios \cite{hentout_2018}.\footnote{For instance, this has been the case in some of our earlier work \cite{mitrevski_2017}.} In addition, exhaustive robot tests need to cover not only the software aspects of the robot system, but also the overall behavior and decisions that the robot makes during task execution \cite{araiza_2016}.

\begin{figure}[tp]
    \centering
    \includegraphics[width=\linewidth]{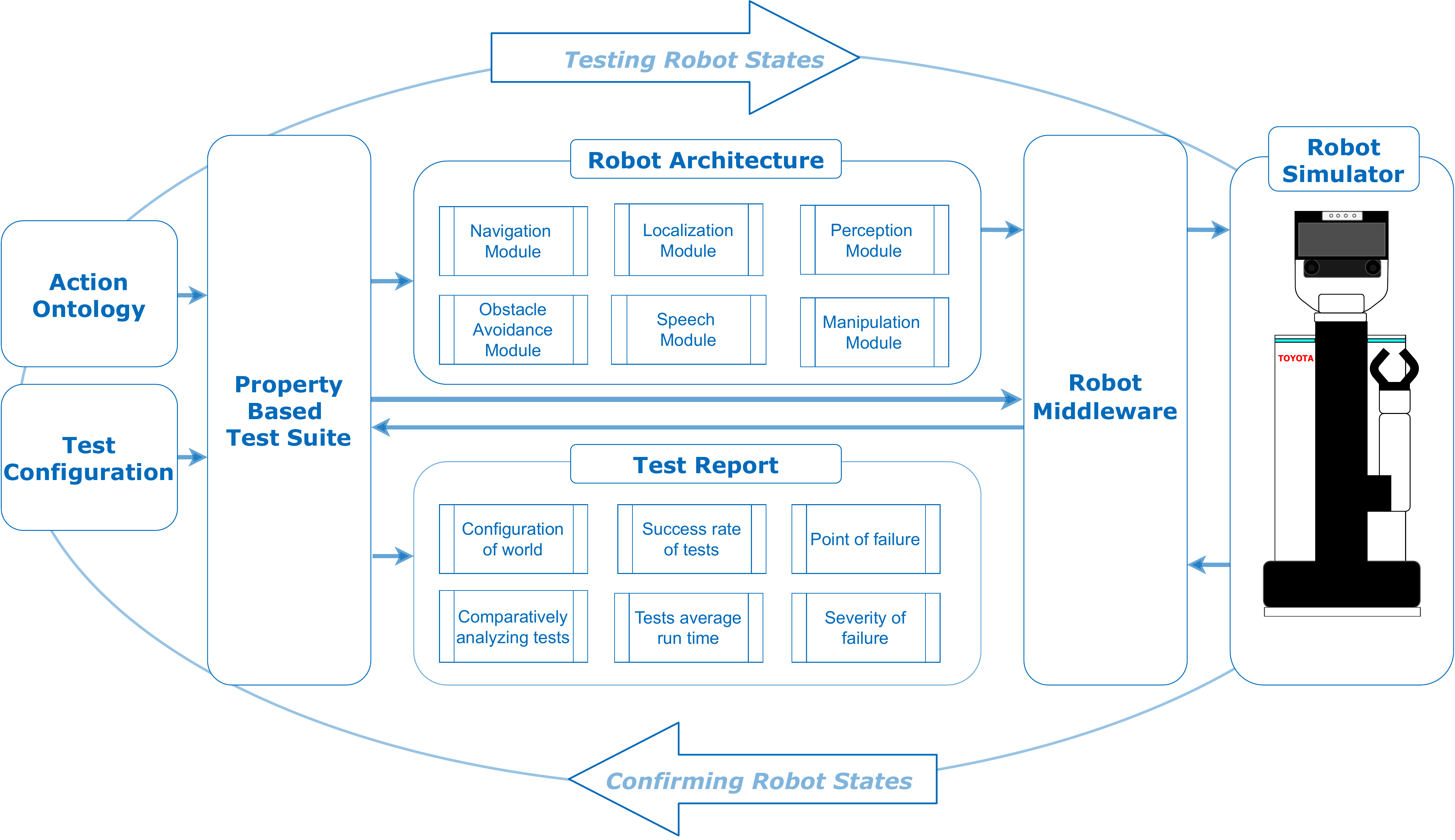}
    \caption{Our proposed simulation-based property-based testing framework for verifying robot action execution}
    \label{fig:overview}
\end{figure}

This paper is motivated by these challenges in robot testing and addresses the question of how to test robot behaviours so that robot execution failures, which go beyond software failures, can be systematically analysed. Our proposed framework uses automatic scenario generation for simulation-based robot testing, namely we generate a set of simulated scenarios in which a robot performs various actions and assesses its performance using property-based testing \cite{george_1997}. The framework includes four core components: (i) an action ontology that relates primitive robot actions to properties of the world that determine the success or failure when those actions are performed, (ii) scenario generation based on which randomised world configurations are created to mimic situations that a robot might encounter, (iii) property-based tests that evaluate the performance of a robot in the generated scenarios, and (iv) report generation that includes the world properties used in the scenario generation as well as the results of the executed property tests, both of which can be used to identify potential reasons for an execution failure. An overview of our framework is provided in Fig. \ref{fig:overview}.
We demonstrate the use of the framework using a Toyota Human Support Robot (HSR) \cite{yamamoto_2019} and our software stack for domestic robots, focusing on functionalities such as point-based navigation, object perception for subsequent manipulation, as well as object pickup and placing in various common domestic object manipulation scenarios.

\section{RELATED WORK}
\label{sec:related_work}

In the context of pure software systems, various testing strategies can be used, such as model-based, functional, mutation, or regression testing \cite{arora_2017}; however, as pointed out by Bihlmaier and W{\"o}rn \cite{bihlmaier_2014}, software testing alone cannot capture the full complexity of robot systems, which interact with the world and other agents. Robot systems thus require dedicated testing methodologies that are able to incorporate physical aspects of the world and take into account the potential of experiencing execution failures.

In the context of simulation-based testing, the core aspect is the scenario generation process.
Scenario generation has been studied fairly thoroughly for autonomous driving.
Majumdar et al. \cite{majumdar_2021} present a test specification language called PARACOSM, which creates realistic environment simulations for testing autonomous driving systems in the Unity physics engine; the language allows defining object and environment properties in a test scenario, thereby providing a declarative way of scenario specification and configuration.
Park et al. \cite{park_2019} propose an approach for simulated scenario generation based on networks trained with real-world driving data; the objective here is to generate realistic driving scenarios that represent multiple vehicle events in a single video session, which enables learning data generation and subsequent vehicle testing.
In \cite{abey_2019}, Bayesian Optimisation is used to generate stimulating adversarial scenarios for an autonomous vehicle; imitation learning is subsequently used to acquire corrective actions from an expert while the scenario is being executed in simulation.
A similar technique is proposed by Koren et al. \cite{koren_2019}, where Adaptive Stress Testing based on a Markov Decision Process is used to identify driving scenarios that are likely to lead to a failure.
Our framework draws upon various insights from the above work. As in \cite{majumdar_2021}, we use a specification language to control various test generation parameters.
Similar to \cite{park_2019}, our scenarios may represent multiple events, but we do not rely on available video data for training a test generator, although it would be possible to integrate the use of video data through an appropriate test generation strategy.
Finally, methods similar to \cite{abey_2019,koren_2019} could be combined with property-based testing by learning to focus on challenging test scenarios, which provides an interesting direction for future work.

In the context of service robots, various testing methodologies have been used.
In \cite{santos_2018}, property-based testing is used for testing robots whose software is based on the Robot Operating System (ROS), such that tests are based on properties about the expected input-output relations represented via the ROS component graph.
Araiza et al. \cite{araiza_2016} compare two test generation strategies - Belief-Desire-Intention and model checking of timed automata - which can be used for testing collaborative service robots.
In \cite{castro_2018}, Estivill-Castro et al. propose a framework for simulation-based robot testing that utilises continuous integration; here, tests are centered around robot behaviours which are modelled by automata, such that the test complexity is governed by a hierarchical structure underlying the behaviours.
As \cite{santos_2018}, our work is based on property-based testing; however, while our robot software is also embedded in ROS, our general approach is independent of a communication framework, namely it allows testing behaviours that do not require a communication framework to be used during execution.
Similar to \cite{araiza_2016}, our test generation strategy is based on modelled relations between robot actions and properties of interest, but our primary focus is on physical interaction with the world and failures that may occur during action execution with a large variety of objects.
As in \cite{castro_2018}, our framework allows robot testing at different levels of complexity, namely we aim to test both individual robot actions as well as complete tasks.

Robot simulations can also be utilised to guide the robot execution process at runtime, which can be seen as an online testing strategy.
Kunze et al. \cite{kunze_2011} propose a simulation-based methodology to identify suitable parameters for executing a robot action by recording logs from simulated executions and analysing those using the event calculus.
Similarly, M{\"o}senlechner and Beetz \cite{mosenlechner_2013} present a methodology to identify a set of valid positions and orientations for a mobile robot's base by considering predictions from simulated events while picking and placing objects.
Our simulated tests are similar to the scenario executions in \cite{kunze_2011,mosenlechner_2013}, but our framework is only intended to be used for comprehensive offline testing of a robot's behavioural components so that failures can be analysed and improved before a robot is practically deployed.


\section{PROPERTY-BASED TESTING IN SIMULATION}

Our proposed framework aims at testing robot actions, such as \emph{pick object} or \emph{move to}, and complete tasks that are composed of multiple actions, using property-based testing in simulation, such that the objective is to automate the test generation and evaluation process so that actions can be tested under different conditions and with different parameters. To make this possible, it is necessary to specify how test scenarios should be generated and which properties should be tested for each action. In this section, we start with a description of property-based testing and explain how we encode and test properties, as well as how tests are documented for subsequent analysis.

\subsection{Property-Based Robot Testing}
\label{sec:property_based_robot_testing}

Property-based testing is a software testing framework that verifies whether certain specifications, or properties, are met in a given software system \cite{george_1997}, such that the idea is to feed a component with varied inputs and check whether it fails with any given input values.\footnote{The mechanism of varied inputs fed into property-based tests is somewhat similar to fuzz testing in which random and possibly invalid data is passed into a software component to detect failures.} Property-based testing is a more general version of unit testing, as a single property-based test can substitute a collection of manually-specified unit tests; in other words, property-based testing can be seen as generative testing that can cover a large amount of the domain space using parameter generation \emph{strategies}.

The general structure for applying property-based testing consists of (i) modelling the expected behavior of a system in terms of properties, (ii) determining the range of parameters that is of interest, (iii) activating an input generator in order to generate parameters in the specified range, and (iv) verifying that the test result satisfies the specified properties.

\begin{figure}[tp]
    \centering
    \includegraphics[width=\linewidth]{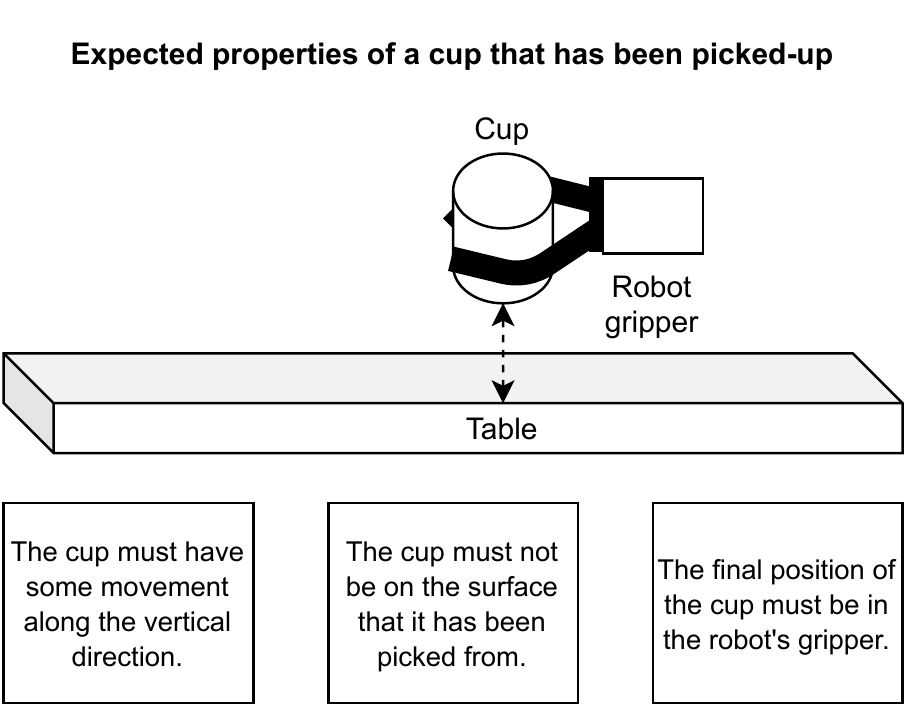}
    \caption{Relevant properties for an object grasping action}
    \label{fig:property_based_test_example}
\end{figure}

An example of applying property-based testing is given by the simple scenario shown in Fig. \ref{fig:property_based_test_example}, in which we want to check whether a robot has picked up an object. The following properties are defined for an object to be picked up: (i) the object must be elevated from the surface that it was picked from, (ii) the object must not be on the surface that it was picked from, and (iii) the object's final position must be in the robot's hand. According to the first property, we expect that a picked up object will always be elevated from its original position on the surface; a violation of this property should result in a failure, as this would indicate that the object was knocked over while the robot was attempting to grasp it or the object was never grasped in the first place. Similarly, the second property identifies whether a successfully grasped object has slipped from the gripper back onto the table during or after the pick action. The third property verifies that the object is in the robot's hand. Through a combination of these three property tests, different types of failures can be identified; for instance, the success of property one and two and failure of property three would mean that the robot successfully picked up the object, but the object did not remain in its hand and fell down on the floor. Defining more properties for a system increases the testing mechanism's ability to identify the point of failure. To increase the coverage for a given use cases, we utilise \emph{input generators}, which pass varied parameters for verifying the system's properties. As described later, these generators are parameterised at test runtime using a minimal configuration.

\begin{figure}[tp]
    \centering
    \includegraphics[width=\linewidth]{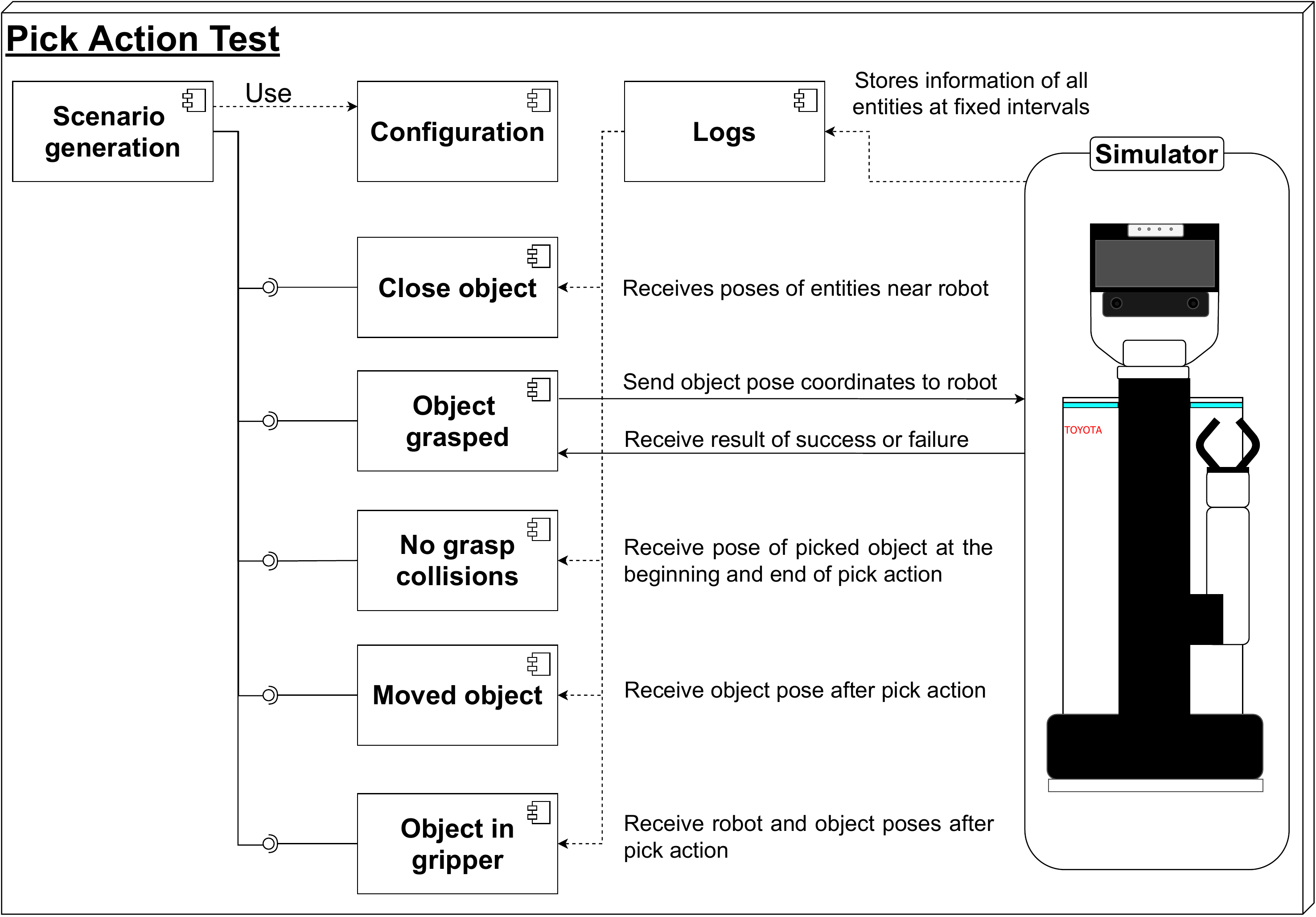}
    \caption{An implementation example of modelling properties for a \emph{pick object} action of a robot.}
    \label{fig:pick_test_implementation_example}
\end{figure}

A more complete definition of relevant properties that we use for testing an object grasping action is given in Table \ref{tab:pickup_properties}; the testing process itself is illustrated in Fig. \ref{fig:pick_test_implementation_example}.

\begin{table}[!h]
    \caption{Defined properties for an object grasping action}
    \label{tab:pickup_properties}
    \begin{tabular}{p{0.13\linewidth} | p{0.75\linewidth}}
    \cellcolor{gray!10!white} \textbf{Property} & \cellcolor{gray!10!white} \textbf{Brief Description} \\\hline
    Close object & Objects should exist within the proximity of the robot before the action\\
    \hline
    Object grasped & The action should report that it has completed its execution \\
    \hline
    No grasp collisions & The positions of all objects, with the exception of the robot and the object to be grasped, should remain the same before and after the action\\
    \hline
    Moved object & The position of the grasped object should change after the action is executed\\
    \hline
    Object in gripper & The grasped object should be in the robot's hand at the end of the action\\
    \hline
    \end{tabular}
\end{table}

Similar properties are also encoded for other actions involved in tabletop manipulation, in particular \emph{move to}, \emph{perceive plane}, and \emph{place object}.

\subsection{Action Ontology}
\label{sec:action_ontology}

We encode the information about which properties to test for a given action in an ontology $\mathcal{O}$ represented in the Web Ontology Language (OWL).\footnote{\url{https://www.w3.org/OWL/}} In the ontology, each action is represented as an instance of an \texttt{Action} class. Actions are performed with respect to a predefined coordinate frame and are associated with multiple properties that need to be tested for verifying the execution success. Properties are modelled as instances of a \texttt{Property} class, where each property depends on parameters that are set in the context of a given action; each such parameter is represented as an instance of a \texttt{PropertyParameter} class, such that the values of the parameters are set at runtime during test generation.

In the ontology, these classes are related to each other through the object properties shown in Table \ref{tab:ontology_properties}.
\begin{table}[htp]
    \centering
    \caption{Ontology properties}
    \label{tab:ontology_properties}
    \begin{tabular}{p{0.3\linewidth} | p{0.2\linewidth} | p{0.34\linewidth}}
        \cellcolor{gray!10!white} \textbf{Object property} & \cellcolor{gray!10!white} \textbf{Domain} & \cellcolor{gray!10!white} \textbf{Range} \\\hline
        \texttt{performedIn}     & \texttt{Action}   & \texttt{CoordinateFrame}   \\\hline
        \texttt{successProperty} & \texttt{Action}   & \texttt{Property}          \\\hline
        \texttt{hasParameter}    & \texttt{Action}   & \texttt{PropertyParameter} \\\hline
        \texttt{needsParameter}  & \texttt{Property} & \texttt{PropertyParameter}
    \end{tabular}
\end{table}
Here, the \texttt{performedIn} property relates an action to the coordinate frame in which the action is performed. \texttt{successProperty} specifies which properties need to be verified for the execution of an action to be considered successful. \texttt{needsParameter} and \texttt{hasParameter} specify which parameters are required by a given property and which parameters are set for testing a given action, respectively. To indicate that the values of property parameters are assigned at runtime, we represent each instance of \texttt{PropertyParameter} by a designator, similar to \cite{winkler2014}. An example ontology model of a \emph{pick object} action, which illustrates what is encoded for each action, is shown below.\footnote{The ontology is available at \url{https://github.com/b-it-bots/action-execution/}}

\begin{mdframed}[style=frameStyle]
    \vspace{-0.25cm}
	\lstset{language=Python,
            basicstyle=\ttfamily\tiny,
            frame=lines,
            breaklines=true,
            showstringspaces=false,
            procnamekeys={def class},
            caption={OWL-based property model of an object grasping action}}
\begin{lstlisting}
<owl:NamedIndividual rdf:about="http://actions#Pick">
    <rdf:type rdf:resource="http://actions#Action"/>

    <action:performedIn rdf:resource="http://actions#BaseLink"/>

    <action:successProperty rdf:resource="http://actions#MovedAlongY"/>
    <action:successProperty rdf:resource="http://actions#OnSurface"/>
    <action:successProperty rdf:resource="http://actions#InHand"/>

    <action:hasParameter rdf:resource="http://actions#objectLocation"/>
    <action:hasParameter rdf:resource="http://actions#objectSurface"/>
    <action:hasParameter rdf:resource="http://actions#pickOject"/>
    <action:hasParameter rdf:resource="http://actions#surfaceObjects"/>
</owl:NamedIndividual>

<owl:NamedIndividual rdf:about="http://actions#OnSurface">
    <rdf:type rdf:resource="http://actions#Property"/>
    <action:needsParameter rdf:resource="http://actions#objectSurface"/>
    <action:needsParameter rdf:resource="http://actions#pickObject"/>
</owl:NamedIndividual>
\end{lstlisting}
\end{mdframed}

Automatic testing of properties is possible by implementing each property $pp$ as a parameterisable function, such that the properties $Props_{a}$ that need to be tested for a given action $a$ are dynamically invoked. Alg. \ref{alg:actionTest} provides a high-level summary of the property testing procedure.

\begin{algorithm}[tp]
    \small
    \begin{algorithmic}[1]
        \Function{\texttt{testAction}}{$a$, $\mathcal{O}$}
            \State $P_a$ \assign \texttt{getActionParameters}($\mathcal{O}$, $a$)
            \State $P_{val}$ \assign $\{ \}$
            \For{$p$ \algin $P_a$}
                \State $p_{val}$ \assign \texttt{generateParameter}($p$)
                \State $P_{val}$ \assign $P_{val}$ $\cup$ $p_{val}$
            \EndFor
            \State \texttt{executeAction}($P_{val}$)
            \State $Props_{a}$ \assign \texttt{getSuccessProperties}($\mathcal{O}$, $a$)
            \For{\texttt{pp} in $Props_{a}$}
                \State $pp_{val}$ \assign \texttt{getPropertyParameters}(\texttt{pp}, $P_{val}$)
                \If{\algnot \texttt{pp}($pp_{val}$)}
                    \State \Return \algfalse
                \EndIf
            \EndFor
            \State \Return \algtrue
        \EndFunction
    \end{algorithmic}
    \caption{Property-based action testing}
    \label{alg:actionTest}
\end{algorithm}

\subsection{Test Scenario Generation}
\label{subsec:scenario_gneration}

Similar to \cite{wen_2020}, we define a test scenario as a combination of an environment configuration and an objective that a robot needs to achieve. The scenario generation process thus requires to (i) set up an environment and (ii) assign a task to the robot being tested. Our scenario generation component serves as a parameterisable input generator which creates scenarios by placing objects in different positions and orientations in a given environment. The generator uses a similar approach to \cite{arnold_2013}; in particular, to generate diverse and dynamic scenarios, we use custom 3D models of different household objects, such that scenarios are generated by placing the models in randomised poses.\footnote{To avoid models from spawning on top of each other due to the randomised nature of the placement, we apply 3D collision prevention using axis-aligned bounding boxes (AABB).} In this paper, we focus on tabletop manipulation scenarios, so objects are generated on surfaces where the robot can manipulate them.

To allow for controlled test coverage, the scenario generation is configured through the parameters in Table \ref{tab:configuration_parameters}.

\begin{table}[htp]
    \centering
    \caption{Scenario configuration parameters}
    \label{tab:configuration_parameters}
    \begin{tabular}{p{0.4\linewidth} | p{0.5\linewidth}}
        \cellcolor{gray!10!white} \textbf{Parameter} & \cellcolor{gray!10!white} \textbf{Description}              \\\hline
        \texttt{tests}                    & List of tests to execute; each test is specified as a list of actions  \\\hline
        \texttt{test\_count}              & Number of times to run each test                                       \\\hline
        \texttt{test\_launcher}           & Path to a launch file that starts all components required by the tests \\\hline
        \texttt{model\_dir}               & Path to a directory of 3D models used in the tests                     \\\hline
        \texttt{worlds}                   & List of possible environments in which a test scenario can take place  \\\hline
        \texttt{model\_list}              & Object models that can be used for manipulation                        \\\hline
        \texttt{nav\_obstacle\_list}      & Object models that can be placed as navigation obstacles               \\\hline
        \texttt{nav\_obstacle\_count}     & Number of navigation obstacles to place in the environment             \\\hline
        \texttt{location\_list}           & Names of locations to which the robot can navigate                     \\\hline
        \texttt{object\_surfaces}         & List of surfaces on which objects can be placed for manipulation       \\\hline
        \texttt{place\_object\_surfaces}  & List of possible surfaces to which objects can be brought
    \end{tabular}
\end{table}

These parameters are passed to the scenario generator through a TOML\footnote{\url{https://github.com/toml-lang/toml}} configuration file. It should be noted that some of the configuration parameters are used to assign the values of property parameters specified in the ontology. This connection is established through TOML tables, where the header specifies the name of the property parameter that should be set and the corresponding configuration parameter defines the list of possible values. The format of the scenario configuration file for a \emph{pick} action test is shown below.

\begin{mdframed}[style=frameStyle]
    \vspace{-0.25cm}
	\lstset{language=Python,
            basicstyle=\ttfamily\tiny,
            frame=lines,
            breaklines=true,
            showstringspaces=false,
            procnamekeys={def class},
            caption={Scenario generation configuration}}
\begin{lstlisting}
tests = [["Pick"]]
test_count = int
test_launcher = str
model_dir = str
nav_obstacle_list = List[str]
nav_obstacle_count = int

[world]
worlds = List[str]

[objectLocation]
location_list = List[str]

[objectSurface]
object_surfaces = List[str]

[pickObject]
model_list = List[str]

[surfaceObjects]
model_list = List[str]
\end{lstlisting}
\end{mdframed}

\subsection{Test Report Generation}
\label{sec:report_generation}

\begin{figure}[tp]
    \centering
    \includegraphics[width=\linewidth]{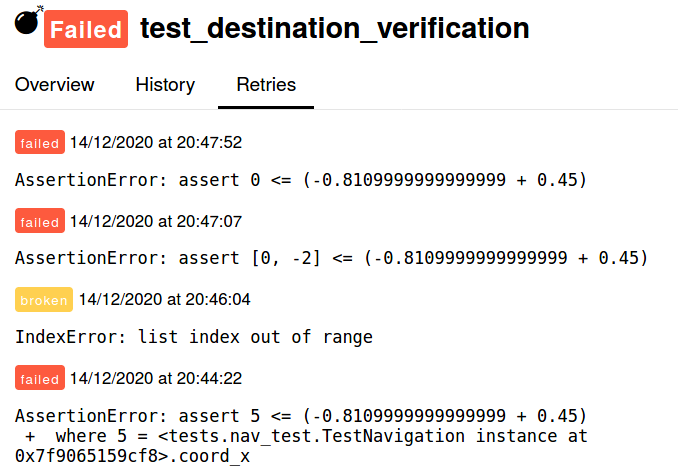}
    \caption{A generated test report for four failed point-to-point navigation runs. In three of these, the navigation goal could not be reached due to a blocking obstacle, while one run failed due to a software error. \textit{Overview} shows information about the overall performance of a test over multiple runs, \textit{History} information about the number of runs of a test, and \textit{Retries} details about the retries of a failed or broken test.}
    \label{fig:test_report}
\end{figure}

For each test, we generate an automatic test report, which is an HTML file created using the \textit{allure} library\footnote{\url{https://docs.qameta.io/allure/}} in conjunction with \textit{Hypothesis} \cite{maciver2019} for property-based testing. The report is generated from json files that are created after each test run, such that it stores information about the test description and duration, world properties and parameters used for the tests, the test consistency and history, as well as the success rate and points of failure of the tests. Test behavior is described by the success rate and consistency of a test over multiple runs; in particular, for each test, the time taken is recorded, such that if the test is inconsistent over multiple runs, the test is considered unstable. Fig. \ref{fig:test_report} illustrates a test report for failed navigation tests.

\subsection{Test Suite Evaluation}
\label{sec:test_evaluation}

We evaluate a test suite, which consists of multiple test runs, through the number of satisfied properties for each involved action over all runs. Let $T$ be a test suite in which $l$ different actions need to be tested. Actions are tested in $m$ randomly generated scenarios $S_k, 1 \leq k \leq m$ with different test parameters; each action $A_{i}^{k}, 1 \leq i \leq l$ has $n$ properties $pp_{j}^{i}, 1 \leq j \leq n$ that are evaluated by an indicator function $\mathbb{I}$. The evaluation of a test suite is thus measured through the normalised success of the individual scenarios:

\begin{equation}
    T = \frac{1}{m}\sum_{k=1}^{m}S_{k}
    \label{eq:test_suite_evaluation}
\end{equation}
where each scenario is evaluated through the involved actions
\begin{equation}
    S_k(A_{1}^{k},\dots, A_{l}^{k}) =  \frac{1}{l}\sum^{l}_{i=1} A_{i}^{k}
    \label{eq:scenario_evaluation}
\end{equation}
and each action is evaluated through its associated properties:
\begin{equation}
    A_i^k(pp_{1}^{i},\dots, pp_{n}^{i}) = \frac{1}{n}\sum^{n}_{j=1}\mathbb{I}_{pp_{j}^{i}=1}
    \label{eq:action_evaluation}
\end{equation}

\section{EXPERIMENTS}
\label{sec:experiments}

To verify our property-based testing approach, we test the Toyota HSR and use Gazebo \cite{gazebo_2004} as a simulation environment since it is directly supported for the HSR.\footnote{The Gazebo setup for the HSR is available at \url{https://github.com/hsr-project}} We use ROS \cite{quigley_2009} as the underlying middleware, such that the tests are targeted at the components in our ROS-based domestic robotics stack.\footnote{The central component of our domestic robotics stack can be found at \url{https://github.com/b-it-bots/mas_domestic_robotics}} ROS is thus used both for inter-component communication as well as for extracting information necessary for executing the property-based tests (such as object poses from Gazebo). We execute all tests on a laptop with an i5-6300HQ CPU at 2.30GHz and 8GB RAM running Ubuntu 16.04 and ROS Kinetic Kame. We present results for two use cases: testing an object grasping action and an object pick-and-place test. For both cases, we report the number of property successes and failures and include cases in which a test run could not be completed due to ROS communication issues\footnote{This particularly refers to action triggering messages that were not received by an action execution component, in which case the robot is waiting for a message indefinitely and thus remains stuck.} during the execution of an action (the latter are referred to as de-synced messages in the figures).

\subsection{Use Case 1: Object Grasping Action}

\begin{figure}[tp]
    \centering
    \begin{subfigure}[t]{0.22\linewidth}
        \centering
        \includegraphics[width=\linewidth]{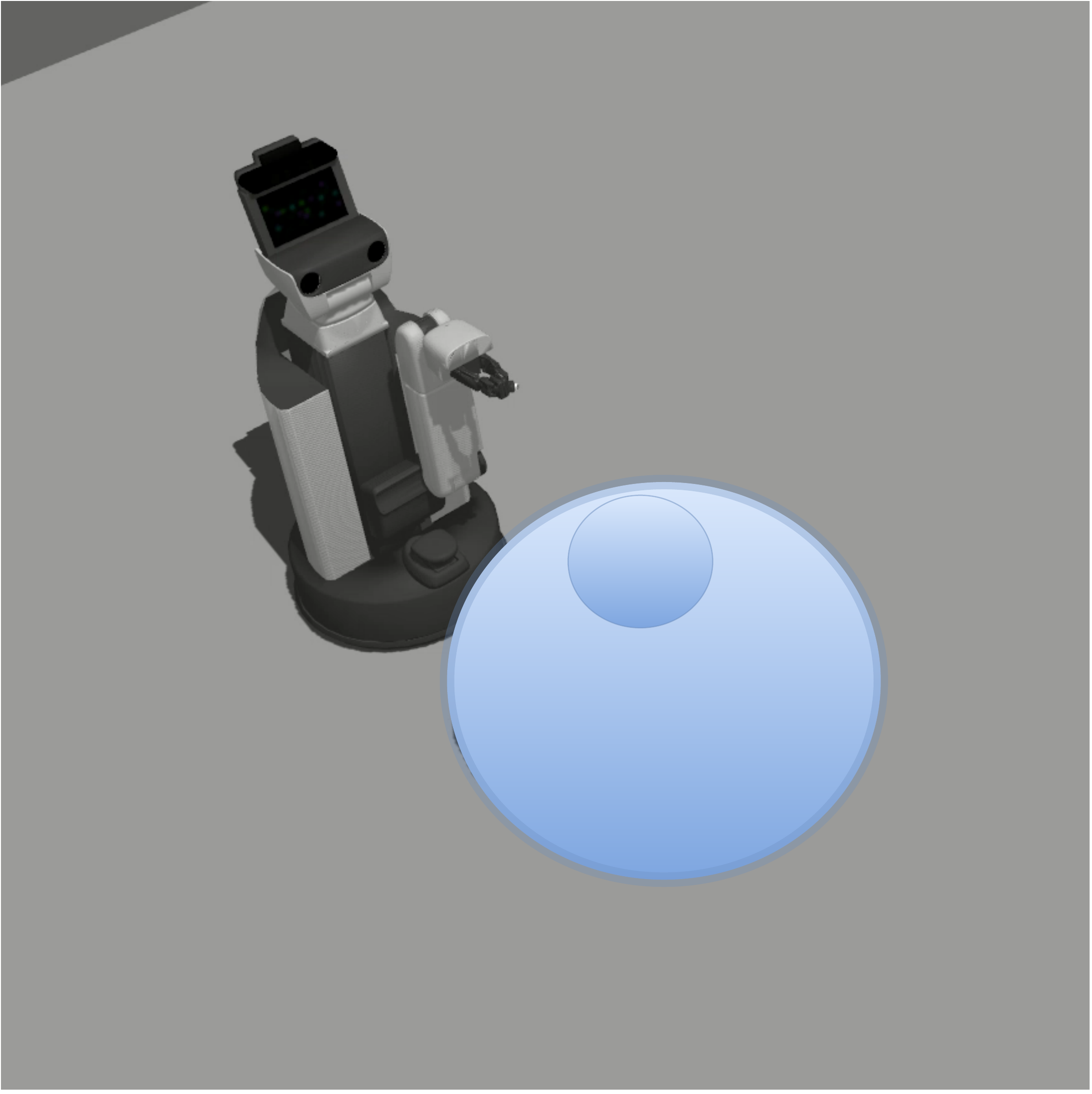}
        \caption{Scenario generation}
    \end{subfigure}
    \hspace{0.01\linewidth}
    \begin{subfigure}[t]{0.22\linewidth}
        \centering
        \includegraphics[width=\linewidth]{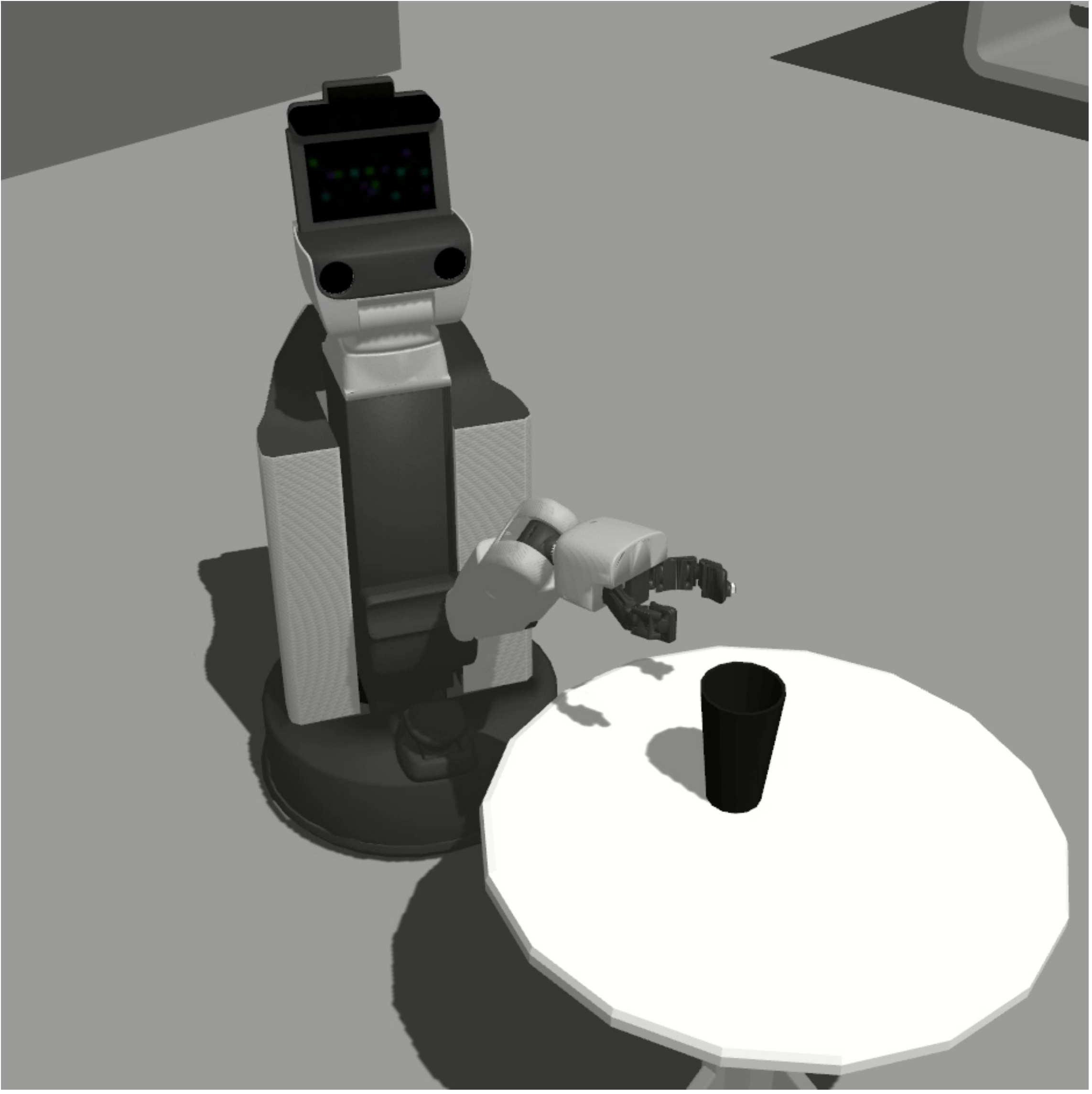}
        \caption{Moving towards the object}
    \end{subfigure}
    \hspace{0.01\linewidth}
    \begin{subfigure}[t]{0.22\linewidth}
        \centering
        \includegraphics[width=\linewidth]{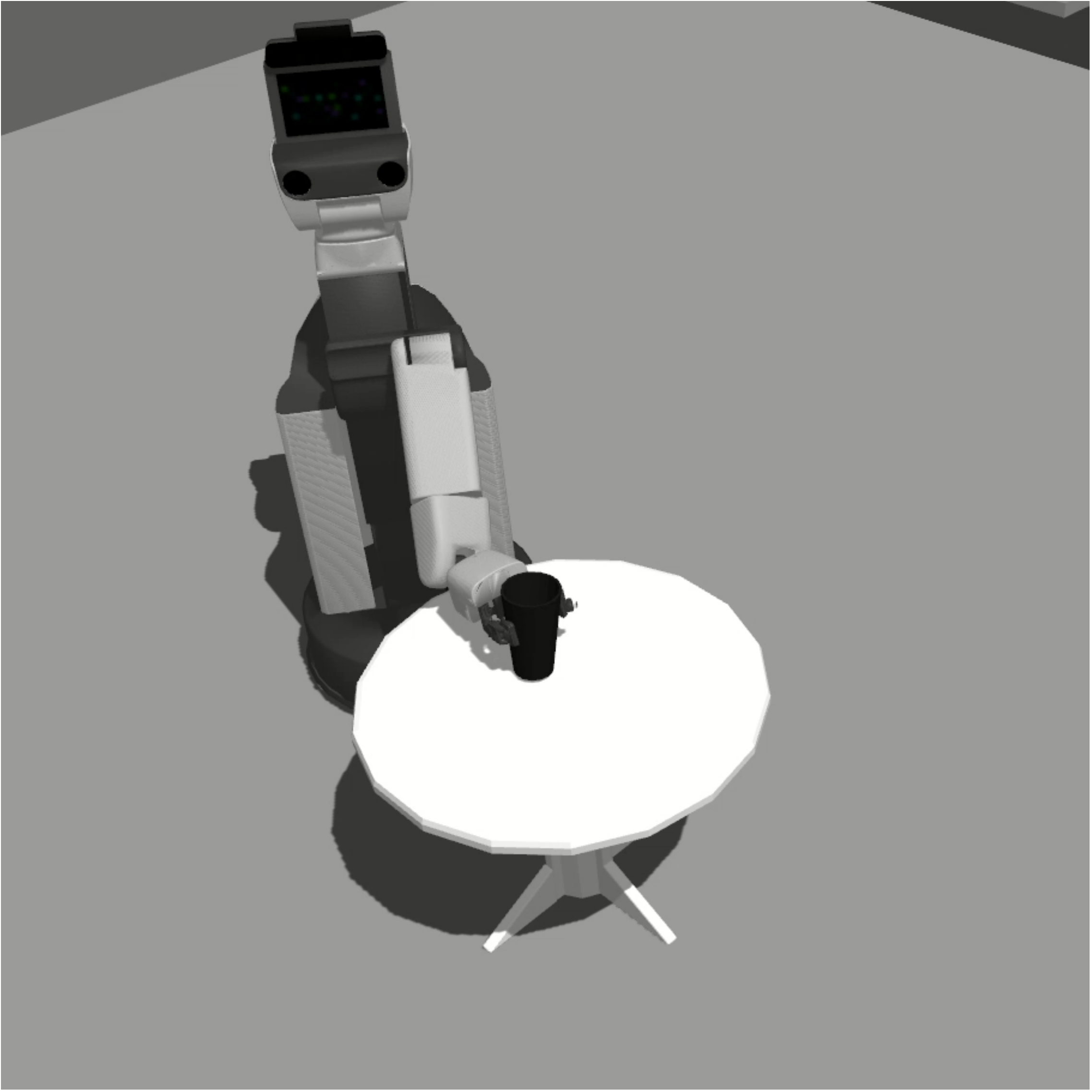}
        \caption{Object approach}
    \end{subfigure}
    \hspace{0.01\linewidth}
    \begin{subfigure}[t]{0.22\linewidth}
        \centering
        \includegraphics[width=\linewidth]{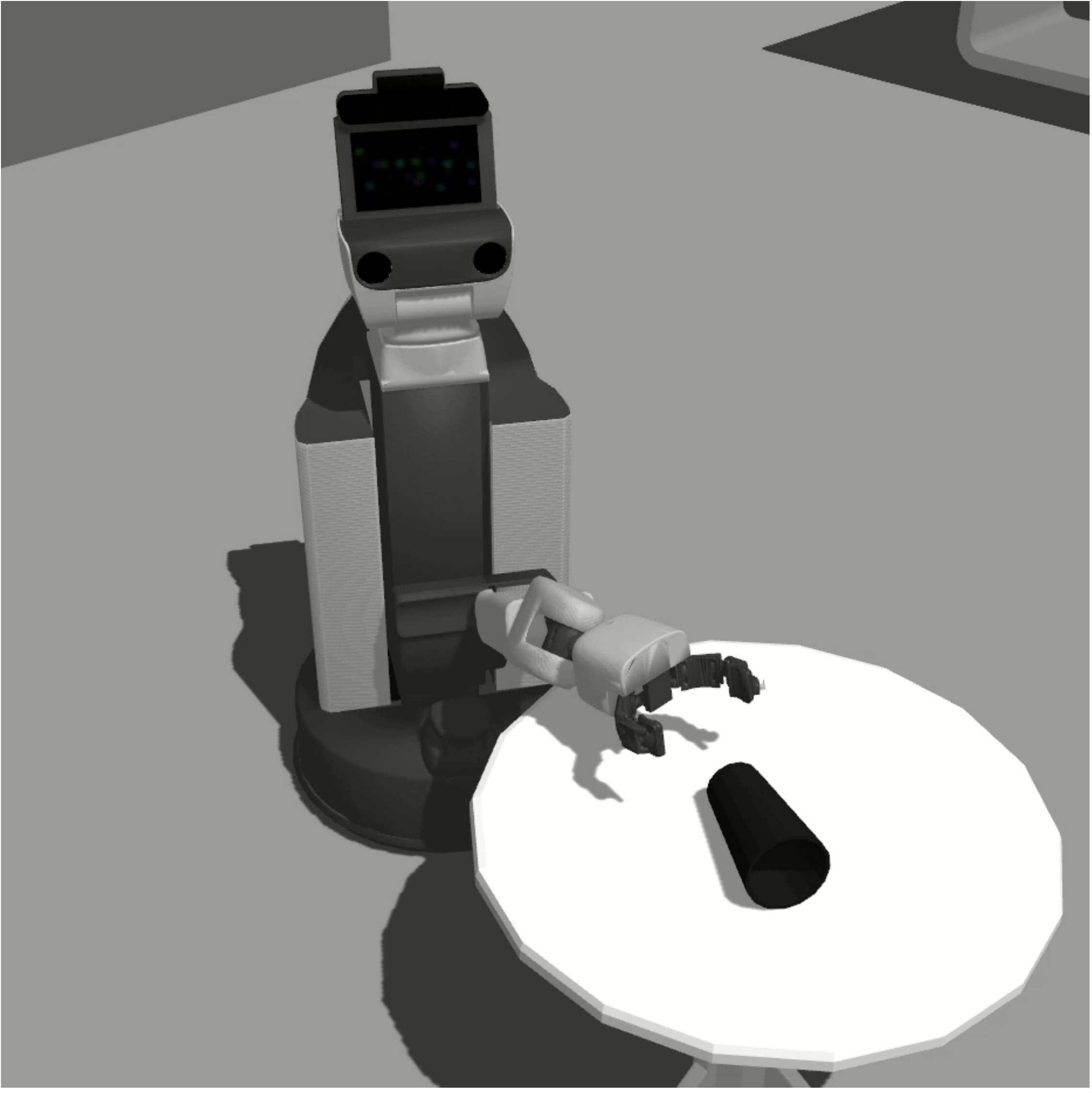}
        \caption{Knocking object over}
    \end{subfigure}
    \caption{Failed grasp execution from a coffee table}
    \label{fig:eval_pick_action_success}
\end{figure}

Our first use case is that of verifying our \emph{pick object} action. Each pick action test scenario follows four main steps, which are illustrated in Fig. \ref{fig:eval_pick_action_success} in the case of a failed grasp.
During scenario generation, an object platform is spawned along with an object on it. For consistency, we performed this test with a small coffee table as an object platform and a glass as the object to be grasped. Once the test scenario is generated, the action is invoked and executed; this is followed by a verification of the properties in Table \ref{tab:pickup_properties}. MoveIt\footnote{\url{https://moveit.ros.org}} is used for trajectory planning and execution in simulation.

\subsection{Use Case 2: Complex Scenario}

\begin{figure}[tp]
    \begin{subfigure}[t]{0.32\linewidth}
        \centering
        \includegraphics[width=\textwidth]{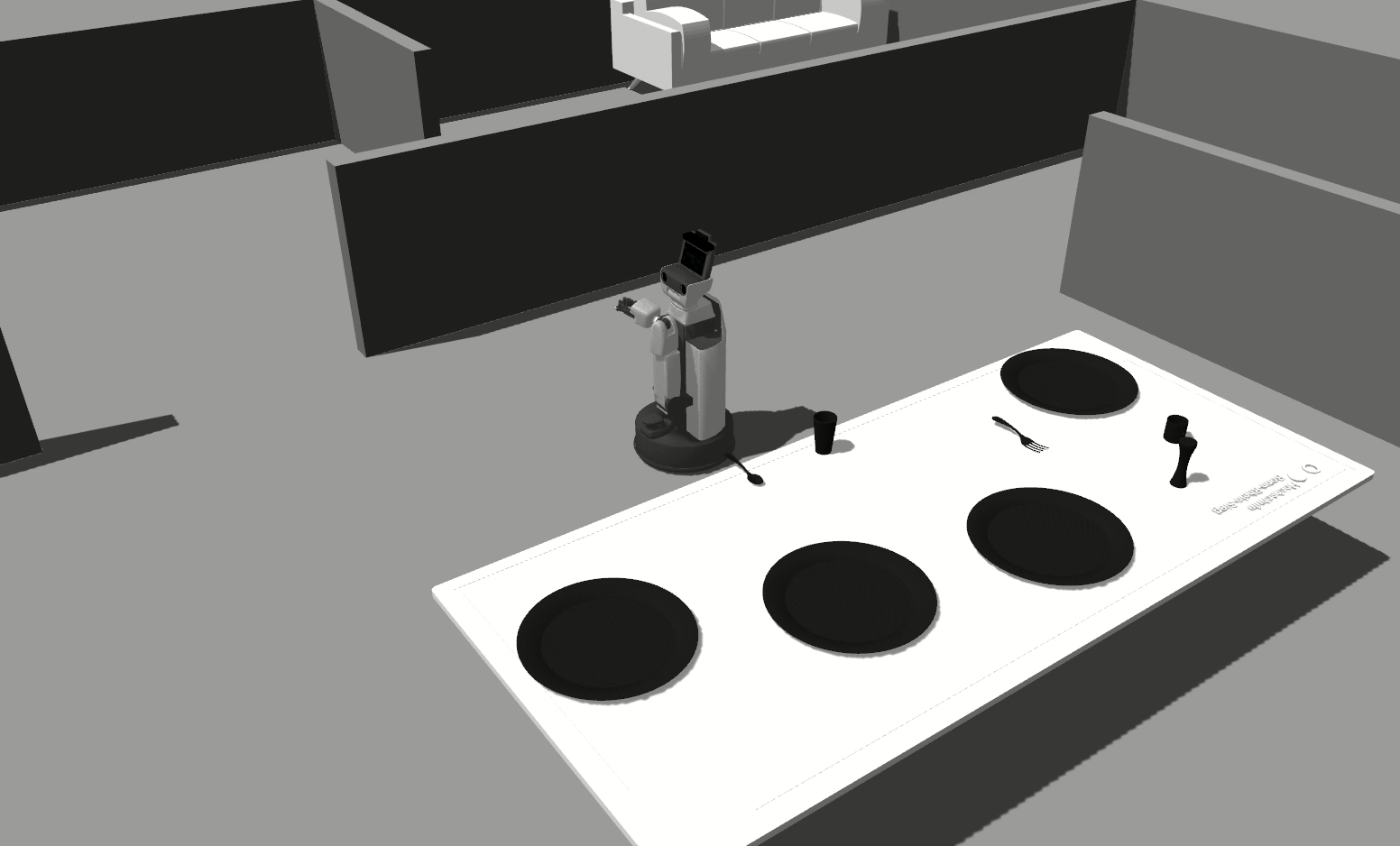}
        \caption{Scenario generation}
    \end{subfigure}
    \hspace{0.05cm}
    \begin{subfigure}[t]{0.32\linewidth}
        \centering
        \includegraphics[width=\textwidth]{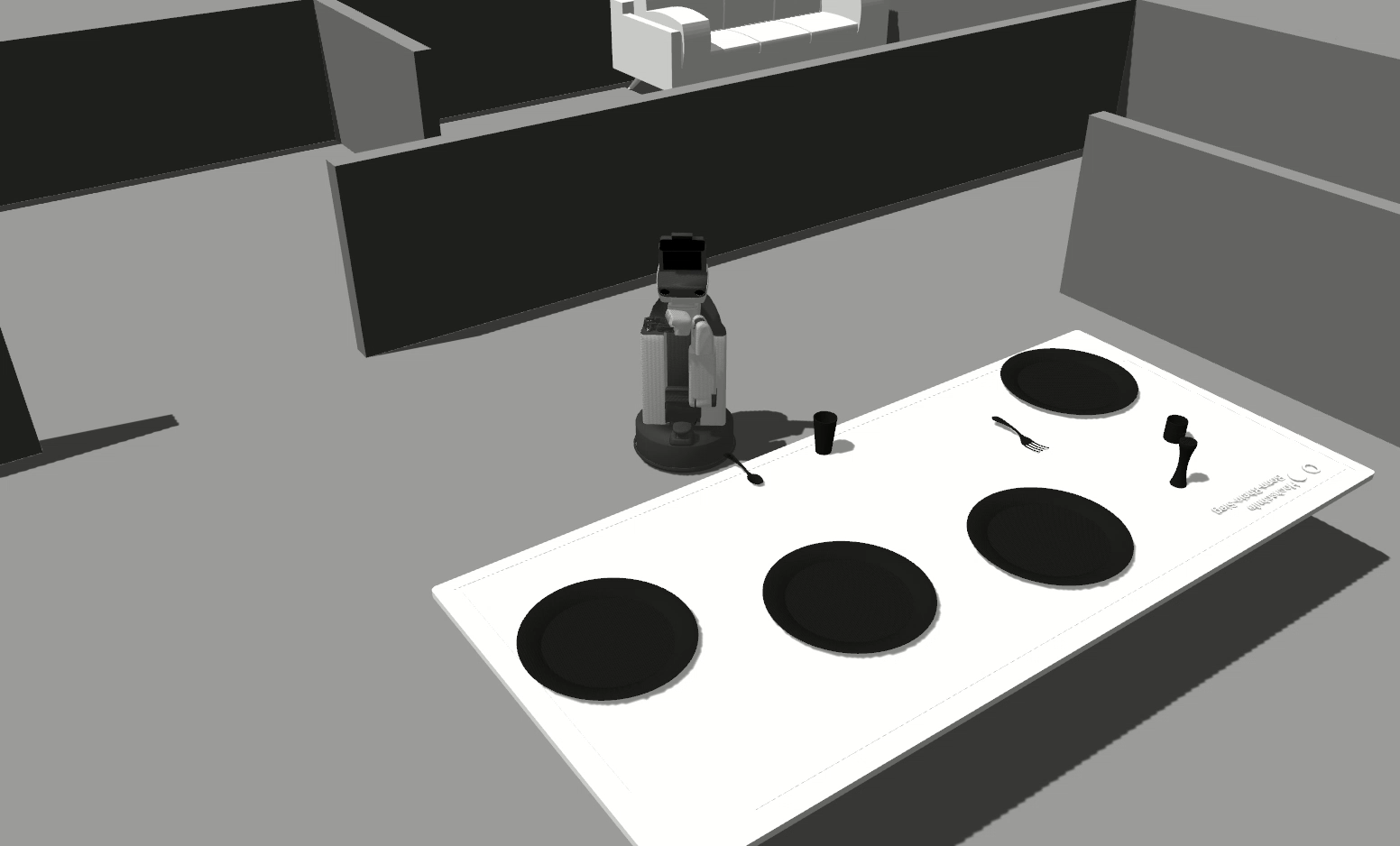}
        \caption{Turning towards the table}
    \end{subfigure}
    \begin{subfigure}[t]{0.32\linewidth}
        \centering
        \includegraphics[width=\textwidth]{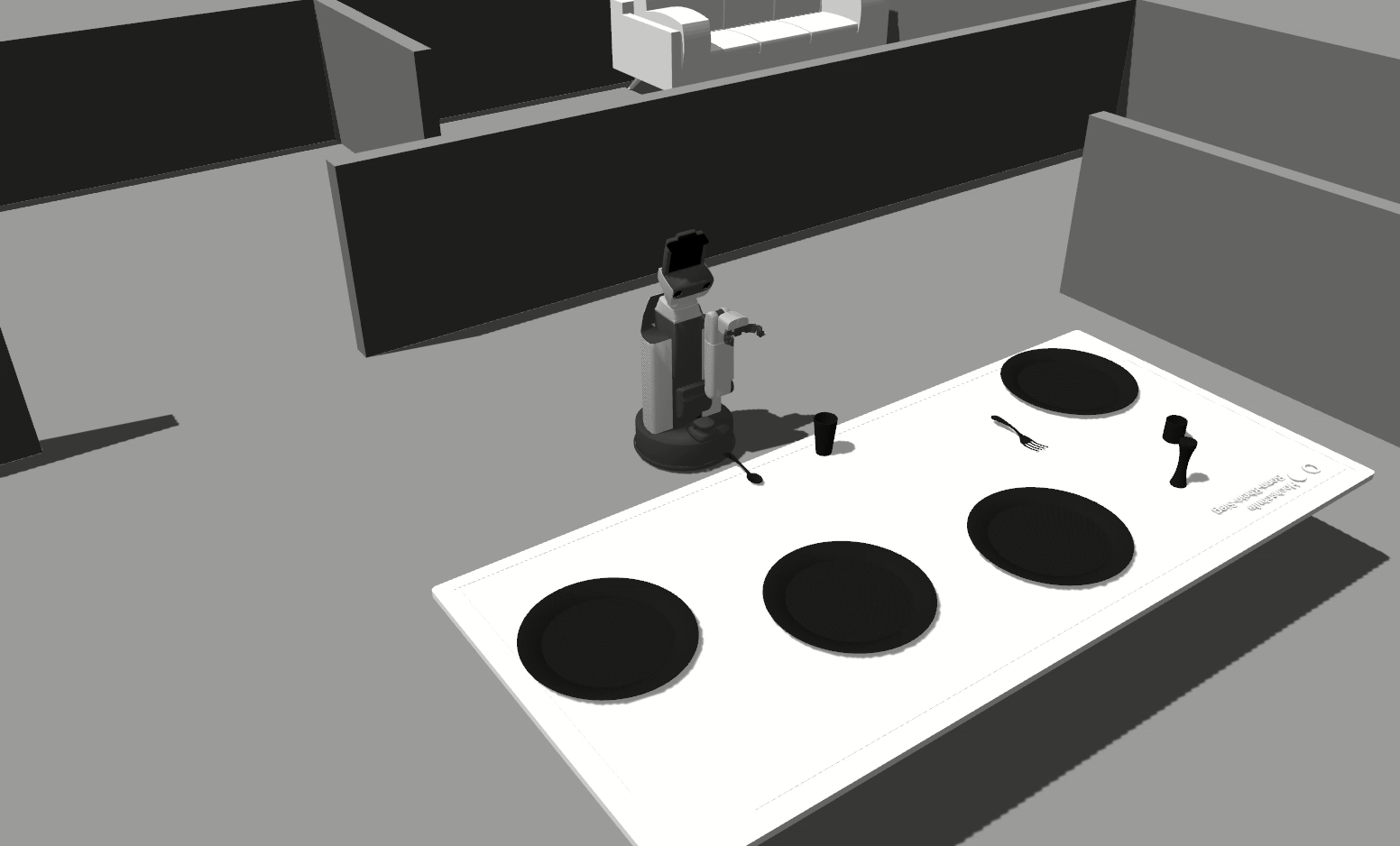}
        \caption{Table and object detection}
    \end{subfigure}
    \newline
    \hspace{0.05cm}
    \begin{subfigure}[t]{0.32\linewidth}
        \centering
        \includegraphics[width=\textwidth]{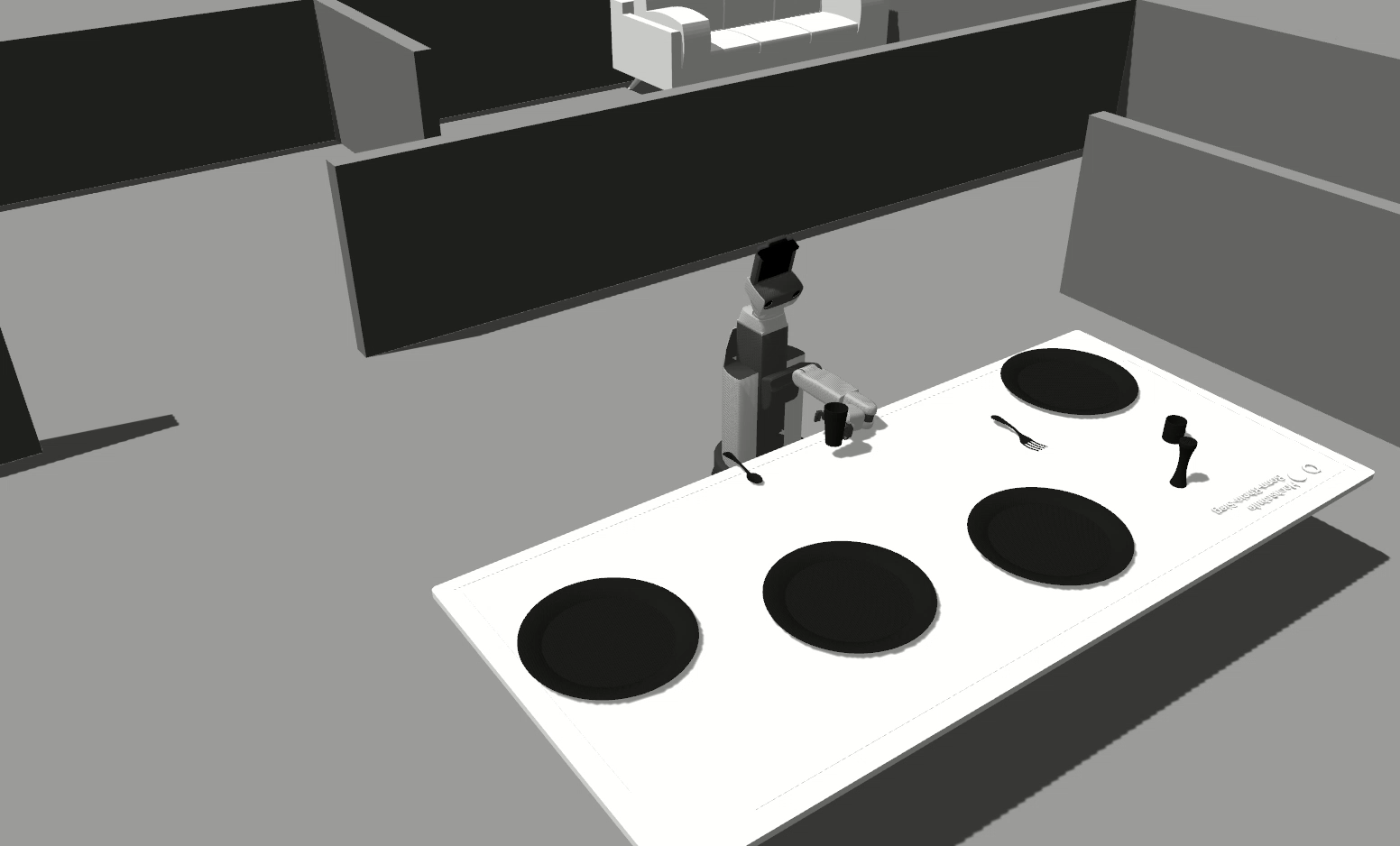}
        \caption{Cup grasping}
    \end{subfigure}
    \begin{subfigure}[t]{0.32\linewidth}
        \centering
        \includegraphics[width=\textwidth]{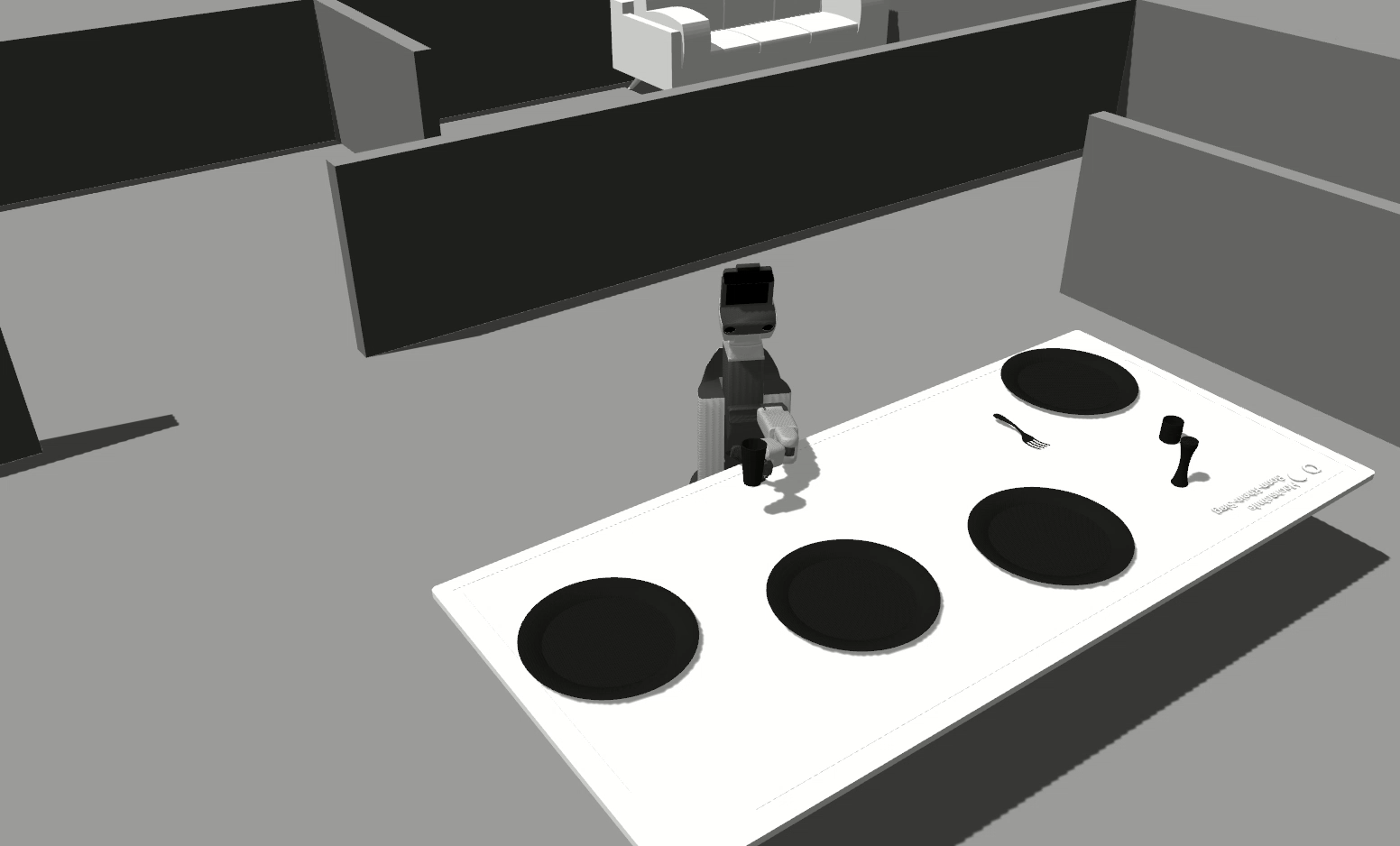}
        \caption{Placing cup}
    \end{subfigure}
    \hspace{0.05cm}
    \begin{subfigure}[t]{0.32\linewidth}
        \centering
        \includegraphics[width=\textwidth]{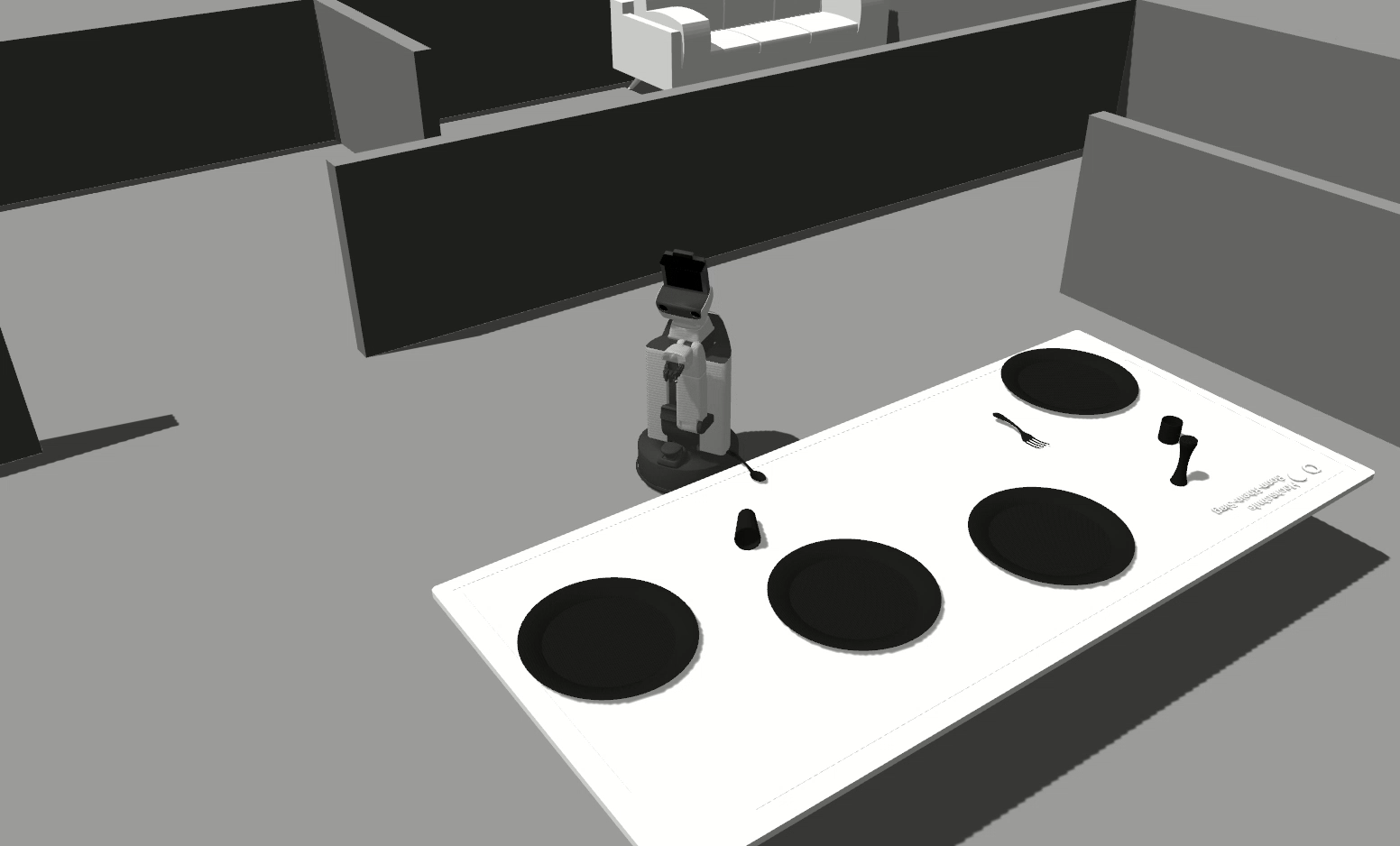}
        \caption{Arm retrieval}
    \end{subfigure}
    \caption{A successful pick-and-place run. Both grasping and placing are performed on the same table.}
    \label{fig:eval_complex_scenario_success}
\end{figure}

The second use case involves a complete object pick-and-place scenario, in which the robot navigates to a table, performs object detection on the table, picks an object, and places it back at a different location on the same table, as illustrated in Fig. \ref{fig:eval_complex_scenario_success}.
The objective of this use case is to validate the robot's behavior when performing a complete task that involves the actions \emph{move to}, \emph{perceive plane}, \emph{pick object}, and \emph{place object}, as the overall success in this task is determined by the sequential success of individual actions, namely task failures are likely to occur due to interdependencies between the actions.

\subsection{Results}

\begin{figure}[tp]
    \centering
    \includegraphics[width=\linewidth]{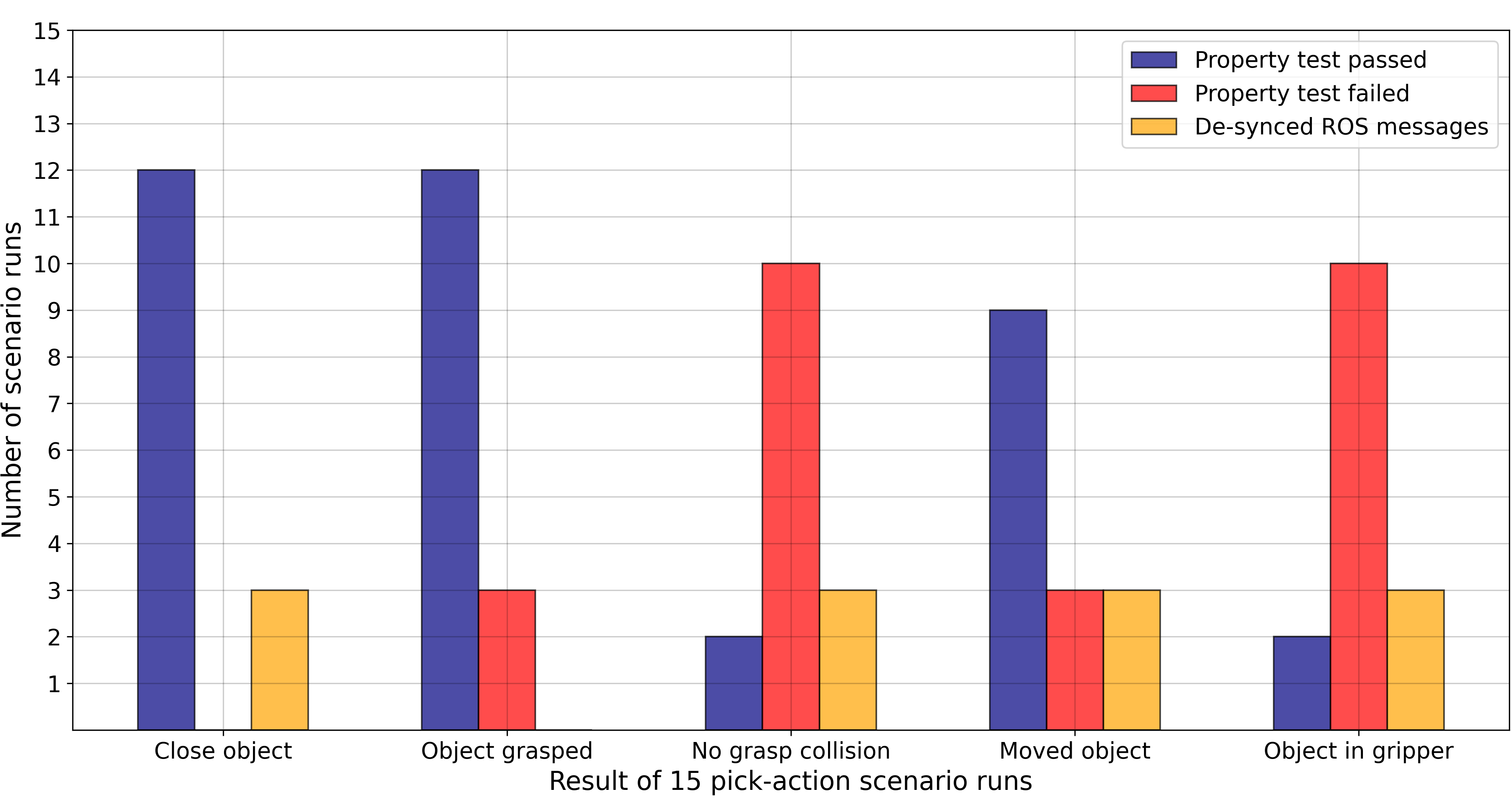}
    \caption{Results of the pick action tests}
    \label{fig:pick_action_evaluation}
\end{figure}

Fig. \ref{fig:pick_action_evaluation} shows the results of 15 runs of the grasping action.
As the results show, the action was fully successful only a few times. Most failures were caused by collisions with the object to be grasped while the approach trajectory was executed; as a result of these collisions, the object was displaced or knocked down and could not be successfully grasped. In some of the runs, the test could not be completed due to lost ROS messages between the components.

\begin{figure}[tp]
    \centering
    \includegraphics[width=\linewidth]{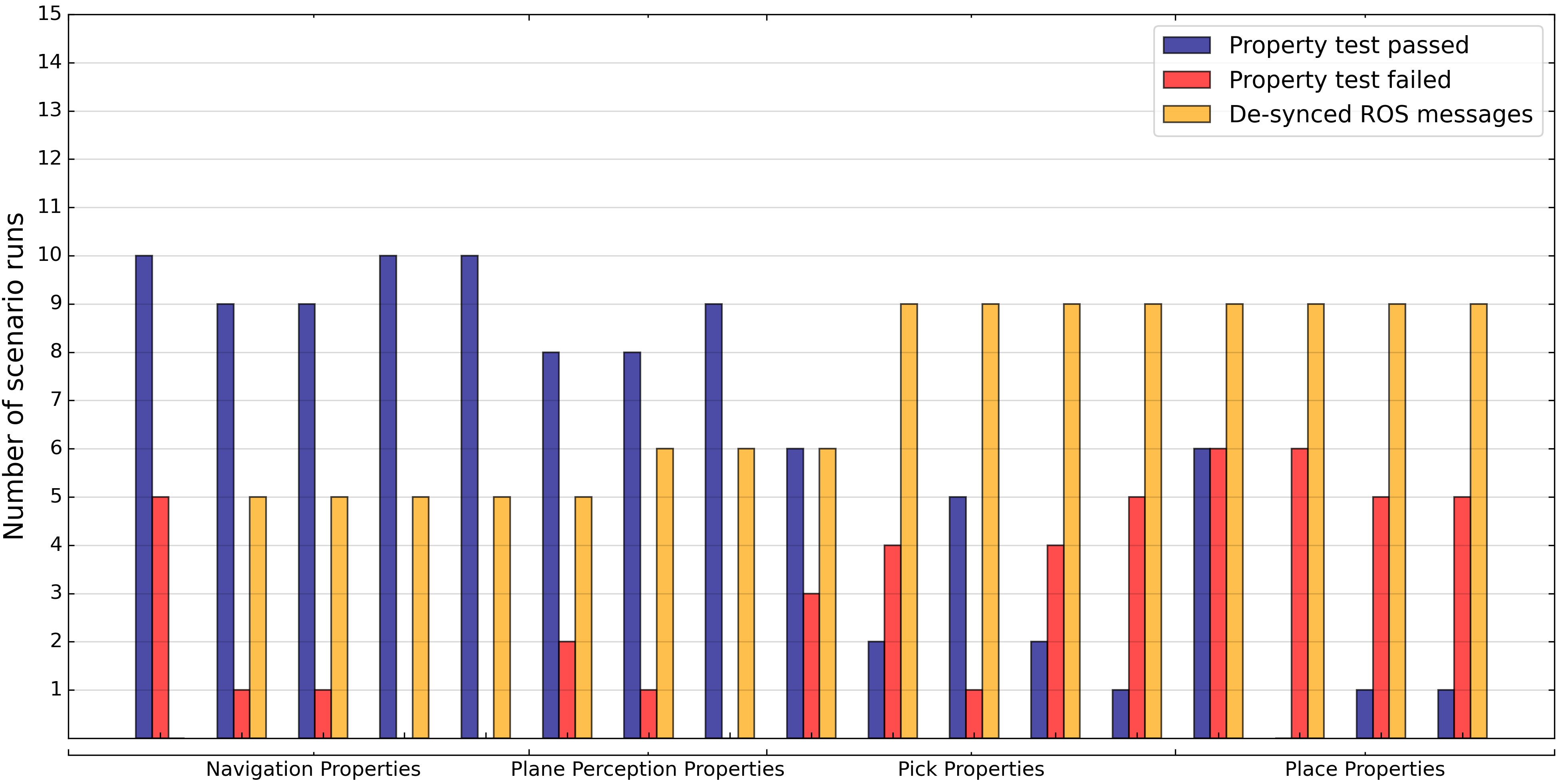}
    \caption{Results of the pick-and-place scenario tests. Properties corresponding to different actions are separated into groups.}
    \label{fig:complex_scenario_evaluation}
\end{figure}

The results of 15 runs of the pick-and-place task are shown in Fig. \ref{fig:complex_scenario_evaluation}.
In this case, the majority of the tests failed due to lost communication messages, particularly during execution of the \emph{pick} and \emph{place} actions. Some of the tests failed while the robot was moving towards the object surface, which was caused by obstacles lying on the designated navigation goal; during execution of the \emph{pick} and \emph{place} actions, a collision was the most common failure cause, just as in the \emph{pick} action test. For completeness, a detailed breakdown of the quantitative evaluation using Eq. \ref{eq:test_suite_evaluation}-\ref{eq:action_evaluation} is provided in Table \ref{tab:complex_scenario_quantitative_evaluation}.

\begin{table}[htp]
    \centering
    \caption{Quantitative evaluation of 15 runs of the pick-and-place scenario. $A_1$-$A_4$ represent the actions \emph{move to}, \emph{perceive plane}, \emph{pick object}, and \emph{place object}, respectively.}
    \label{tab:complex_scenario_quantitative_evaluation}
    \begin{tabular}{p{0.075\linewidth} | p{0.1\linewidth} | p{0.1\linewidth} | p{0.1\linewidth} | p{0.1\linewidth} | p{0.1\linewidth}}
        \cellcolor{gray!10!white} \textbf{Run} & \cellcolor{gray!10!white} $A_1^k$ & \cellcolor{gray!10!white} $A_2^k$ & \cellcolor{gray!10!white} $A_3^k$ & \cellcolor{gray!10!white} $A_4^k$ & \cellcolor{gray!10!white} $S_k$ \\\hline
        1  & 1.0 & 1.0  & 0.4 & 0.25 & 0.67 \\\hline
        2  & 1.0 & 0.67 & 0.0 & 0.0  & 0.42 \\\hline
        3  & 0.6 & 0.33 & 0.6 & 0.25 & 0.45 \\\hline
        4  & 0.0 & 0.0  & 0.0 & 0.0  & 0.0 \\\hline
        5  & 1.0 & 1.0  & 0.0 & 0.0  & 0.5 \\\hline
        6  & 0.0 & 0.0  & 0.0 & 0.0  & 0.0 \\\hline
        7  & 0.0 & 0.0  & 0.0 & 0.0  & 0.0 \\\hline
        8  & 1.0 & 1.0  & 0.4 & 0.25 & 0.67 \\\hline
        9  & 1.0 & 1.0  & 0.4 & 0.25 & 0.67 \\\hline
        10 & 0.0 & 0.0  & 0.0 & 0.0  & 0.0 \\\hline
        11 & 1.0 & 1.0  & 0.4 & 0.25 & 0.67 \\\hline
        12 & 1.0 & 1.0  & 1.0 & 0.75 & 0.94 \\\hline
        13 & 0.0 & 0.0  & 0.0 & 0.0  & 0.0 \\\hline
        14 & 1.0 & 1.0  & 0.0 & 0.0  & 0.5 \\\hline
        15 & 0.0 & 0.0  & 0.0 & 0.0  & 0.0 \\\hline
        \multicolumn{5}{r|}{\cellcolor{gray!10!white} $T$} & \cellcolor{gray!10!white} 0.37
    \end{tabular}
\end{table}

It should be mentioned that the objective of the experiments is evaluating the effectiveness of our proposed approach rather than our components as such.\footnote{Particularly since some of the components that we use on the real robot have not been adapted to the simulation.} From this point of view, configurable property-based testing is able to consistently identify failed actions and generates useful information for finding the likely causes of execution failures.

\section{DISCUSSION AND CONCLUSIONS}
\label{sec:discussion}

In this paper, we proposed a framework for property-based testing of robots in simulation, with a particular focus on tabletop manipulation for a domestic robot. The framework utilises (i) a simulator that allows scenario randomisation and has information about the absolute state of the world, (ii) an ontology for encoding information about relevant action properties and parameters associated with those, and (iii) a configurable scenario generation component. The combination of these components allows testing robot actions, as well as action sequences, in diverse world configurations, which results in test reports that can be used to identify common failure patterns. In experiments with an object grasping action and a pick-and-place task, we showed that property-based testing in simulation is a promising approach that can potentially increase the transparency of service robots and provides insights about the reliability of tested components.

The work in this paper is only a preliminary step towards the use of property-based tests in simulation for regular robot testing. As can be seen from the experimental results, a considerable number of tests failed due to unsuccessful component communication through ROS; we have pinpointed this issue to limitations of the hardware used for running the tests, namely the resource-intensiveness of the simulation was negatively affecting the performance of the complete system, which means that more powerful hardware should be used for running tests so that they are more representative of the real system's behaviour. The Gazebo simulator, which we used for running the tests, was another frequent cause of failures; in particular, Gazebo crashes prevented several test runs from completing successfully, so it would be worthwhile to consider using alternative simulators instead \cite{collins2021}. Another limitation of our current approach is that some of the components that we use in simulation, such as the trajectory execution component, are modified versions of those used on the real robot due to minor discrepancies between the available interfaces on the real robot and in simulation; for the tests to correspond to the real system more closely, the interfaces in simulation and on the robot should be made as close as possible to each other. The realism of the tests is another potential concern for the proposed approach; this is particularly the case in the context of object interaction \cite{collins2019}, although that would not be a significant problem if the simulated tests are only used as a preliminary step before running more critical tests on a real system and, depending on the tested policies, direct transfer to the real world may be possible in certain cases, as for instance shown in \cite{hermann2020}.

Future work will focus on testing execution policies that have been learned with real-world data, particularly those in \cite{mitrevski_2020}, in diverse scenarios. Another aspect that should be addressed is that of improving the scenario generation and expanding the framework so that scenarios other than tabletop manipulation can be tested; this includes incorporating tests in which action sequences are generated by a task planner rather than being predefined during scenario generation. Adding additional scenario configuration parameters, particularly ones that control the physical properties of the objects used in the scenarios, would also contribute to the extensiveness and realism of the tests. Finally, it would also be useful to investigate post-processing of the generated test results so that the data can be utilised for simulation-based learning and policy improvement.


\addtolength{\textheight}{-12cm}   


\bibliographystyle{IEEEtran}
\bibliography{references}

\end{document}